\documentclass{article}
\usepackage{arxiv}

\usepackage{xcolor}
\usepackage[utf8]{inputenc} 
\usepackage[T1]{fontenc}    
\usepackage{hyperref}       
\usepackage{url}            
\usepackage{booktabs}       
\usepackage{amsfonts}       
\usepackage{nicefrac}       
\usepackage{microtype}      
\usepackage{lipsum}
\usepackage{fancyhdr}       
\usepackage{graphicx}       
\usepackage{algorithm}
\usepackage{algpseudocode}
\usepackage{algcompatible}
\usepackage{wrapfig}

\usepackage{xspace}
\usepackage{soul}
\usepackage{mathtools}
\usepackage{makecell}
\usepackage{nicefrac}

\newcommand{\la}{\leftarrow}

\newcommand{\sut}{\text{s.\,t.\,}}

\definecolor{dgreen}{rgb}{0.0, 0.5, 0.0}

\usepackage{cleveref}

\title{Combining model-predictive control and predictive reinforcement learning for stable quadrupedal robot locomotion}
\author{Vyacheslav Kovalev, Anna Shkromada, Henni Ouerdane and Pavel Osinenko}

\begin{document}
\maketitle
\begin{abstract}
Stable gait generation is a crucial problem for legged robot locomotion as this impacts other critical performance factors such as, e.g. mobility over an uneven terrain and power consumption. Gait generation stability results from the efficient control of the interaction between the legged robot's body and the environment where it moves. Here, we study how this can be achieved by a combination of model-predictive and predictive reinforcement learning controllers. Model-predictive control (MPC) is a well-established method that does not utilize any online learning (except for some adaptive variations) as it provides a convenient interface for state constraints management. Reinforcement learning (RL), in contrast, relies on adaptation based on pure experience. In its bare-bone variants, RL is not always suitable for robots due to their high complexity and expensive simulation/experimentation. In this work, we combine both control methods to address the quadrupedal robot stable gate generation problem. The hybrid approach that we develop and apply uses a cost roll-out algorithm with a tail cost in the form of a Q-function modeled by a neural network; this allows to alleviate the computational complexity, which grows exponentially with the prediction horizon in a purely MPC approach. We demonstrate that our RL gait controller achieves stable locomotion at short horizons, where a nominal MP controller fails. Further, our controller is capable of live operation, meaning that it does not require previous training. Our results suggest that the hybridization of MPC with RL, as presented here, is beneficial to achieve a good balance between online control capabilities and computational complexity.

\end{abstract}

\keywords{Legged robots, quadrupedal robots, reinforcement learning, model-predictive control}


\section{Introduction}
\label{sec:introduction}

Modern quadrupedal robots are highly praised for their high degree of mobility, maneuverability, and ability to traverse through diverse terrains, making them well-suited for applications such as inspection and delivery \cite{bellicoso2018advances,biswal2021development}.
However, these robots' performance comes at the price of complex mechanical structures with many degrees of freedom to control.
As a result, developing control systems that enable these robots to efficiently move and operate in dynamically changing environments is one of the most important tasks in the field of quadruped robotics research.

Model predictive control (MPC) includes a range of widely adopted control methods \cite{Camacho2007} not only capable of efficiently handling industrial problems that entail complex processes \cite{Soest2006,Kouro2009,Ma2012}, but that has also been successfully applied to other problems such as indoor microclimate control given its ability to operate over finite prediction horizons \cite{Castilla2014,Scherer2014,Sturzenegger2016,Ryzhov2019}.
In one of the early works on MPC's application to the motion of a real quadrupedal robot \cite{cheetahConvexMPC}, the nonlinear optimization problem was converted into a quadratic program (QP) by linearizing the system dynamics along a desired walking trajectory.
Importantly, the accuracy of the immediate robot dynamics approximation turned out to be more significant than that of the robot dynamics approximation over the prediction horizon.
Notwithstanding the advantages of linearization of the system dynamics, nonlinear MPC finds application for quadrupedal robots \cite{Quad-SDK}.

As traditional MPC solvers can suffer from short planning horizon, convergence to local optima, errors in the dynamics model, and failure to account for future replanning \cite{Jain2021OptimalCD}. 
Several new MPC-based control schemes have been introduced, each aiming to improve specific characteristics of interest. For instance, the integration of classic MPC and whole-body control (WBC) as described in \cite{WBIC} enables the achievement of high-speed locomotion. 
The issue of short prediction horizon, which arises from the requirements to computational complexity, can potentially be solved by incorporating learning based methods due to their ability to approximate objective parameters. \cite{Karnchanachari2020}.
And to provide stability guarantees, a fusion of control Lyapunov function (CLF) with MPC was proposed \cite{CLF-MPC}. 

Reinforcement Learning (RL) has been applied in game applications such as \cite{Akkaya2019,Silver2016,Silver2018,Vinyals2019}, and it finds an application for managing complex problems such as quadrupedal robot control.
The framework proposed in \cite{ji2022concurrent} trains both the state estimator and the control policy concurrently, enabling the fastest reported legged locomotion through the use of RL.
In \cite{lee2020learning}, the policy is learned to enable the robot to move across a challenging terrain, thus showing that RL can successfully manage environmental variability. 
Additionally, an RL method capable of adjusting to the shifts in the robot dynamics was proposed in \cite{hwangbo2019learning}.

To alleviate the short horizon problem, two general approaches incorporating RL in MPC currently exist.
One approach targets the horizon length directly by learning its optimal value, while in the other the need for more prediction intervals is implicitly accounted for when optimizing the cost function.
In the first group of methods, the optimal horizon length can be determined either online \cite{Bohn2021, Bohn2023} or from previously collected data \cite{Gardezi2018MachineLB}.
Both online and offline methods guarantee the prediction horizon to be optimal over the state space; however, to satisfy the real-time performance requirements, the online methods require the usage of simple models, whereas using the pre-collected data, though less computationally demanding, warrants a properly collected training dataset. The motivation behind the second group of methods is the fact that if the terminal cost is accurate enough with respect to the infinite horizon solution to the optimization problem, the controller will demonstrate decent performance even with a shorter prediction horizon \cite{LowreyRajeswaranKakadeTodorovMordatch2019,Zhong2013}.
Hence, it is reasonable to obtain an optimal terminal cost function with learning. An example of such implementations is the so-called roll-out RL schemes (RQL) \cite{osinenko2021effects}.
RQL can be regarded as an extension or enhancement of the traditional MPC.
Note that several works also reported that learning the total cost, either online \cite{Jain2021OptimalCD} or offline \cite{Sawant2023LearningbasedMF, Abdufattokhov2021}, can also be beneficial in case of inaccurate model dynamics and replanning.

Motivated by the good results that RQL demonstrated on simple models \cite{Dobriborsci2021}, we analyze the potential of this method when applied to problems involving more complex systems, such as quadrupedal robots locomotion.
For that purpose, we investigate the RQL performance in comparison to traditional MPC as its baseline considering an experimental setup with an A1 Unitree quadrupedal robot.
By examining the similarities and differences between RQL and MPC, we thus shed light on the potential improvements and performance gains that can be achieved through the integration of RL and MPC techniques in the realm of robotics control.
Our primary observation is that, remarkably, RQL outperformed the baseline MPC in terms of accumulated running cost on the short prediction horizon, thus suggesting that stacked approaches may be viable alternatives to plain RL for quadrupedal robots.

The article is organized as follows.
A description of the mathematical model of the system's dynamics in its environment is given in \cref{sec:Environment}.
\Cref{sec:methods} outlines the control goal and methodology using MPC and RQL methods.
The experimental setup of the robot A1 Unitree is presented in \cref{sec:experiment}.
The article ends with concluding remarks in \cref{sec:results}.

\textbf{Notation}. 
Any vector with a $k$ subscript (e.g. $v_k$) represents the vector state at time step $k$.
For any vector such notation $\{u_{i|k}\}^N_i$ means sequence of vectors $u_i$ for $i = 1 \ldots N$, for time step $k$.

\section{System's dynamics}
\label{sec:Environment}
\begin{figure}[t!]
\includegraphics[width=0.7\textwidth]{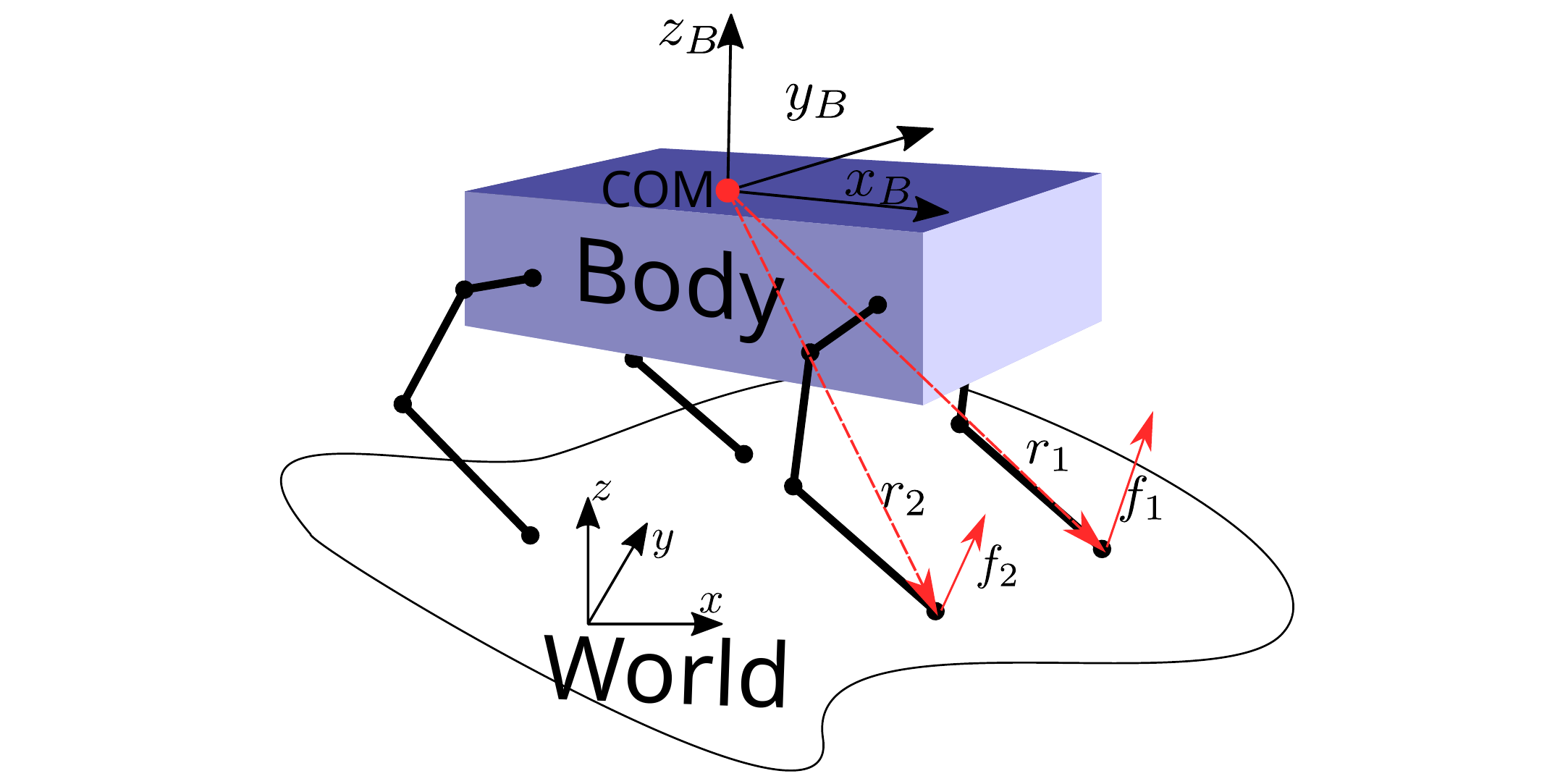}
\centering
\caption{\textbf{World and body coordinate systems. The variables $r_i$ and $f_i$ represent the levers and ground reaction forces.}\label{fig:robot}}
\end{figure}


\begin{figure}[t!]
\includegraphics[width=0.7\textwidth]{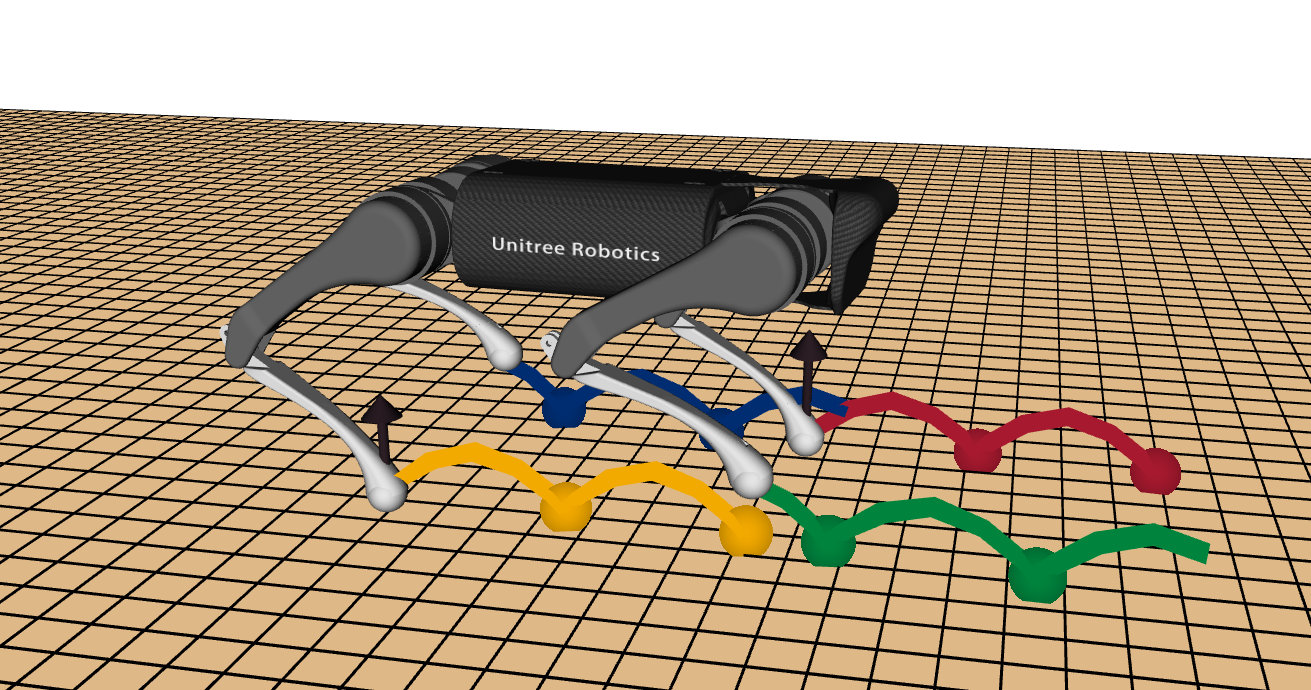}
\centering
\caption{\textbf{Robot simulation (Unitree A1). Black arrows represent ground reaction forces. Colored points and lines represent feet trajectory.}\label{fig:robot_sim}}
\end{figure}

In this work, we take the Unitree A1 robot model as the environment baseline. The system dynamics formulation is inspired by \cite{cheetahConvexMPC,WBIC}, where the robot is considered as a single rigid body under the action of forces at the contact patches, and leg dynamics are neglected due to the small mass ratio of the legs to the main body.
This approximation is reasonable because the total mass of the legs (except the hip joint, which is considered part of the main body) is roughly about $10\%$ of the total mass. The rigid body dynamics in world coordinates are given by \cite{cheetahConvexMPC}:
\begin{align}
\ddot{p}   & = \sum^{4}_{i=1}{\frac{f_i}{m}}-g,  \\
\label{eq:mom_integral}
\frac{\text{d}}{\text{d}t}(\mathcal{I}\omega) & = \sum^{4}_{i=1}{r_i \times f_i}, \\
\dot{R}         & = \omega \times R,
\end{align}

\noindent where $\ddot{p}$ is the second time derivative of the robot position $p$, $f_i$ the $i$-th ground reaction force with corresponding levers $r_i$ (see Figure \ref{fig:robot}), $m$ the total robot mass, $g$ the gravitational acceleration, $\mathcal{I}$ the moment of inertia, $R$ the rotation matrix and $\omega$ the angular velocity. 

The robot orientation is determined using Euler angles: $\Theta = [\phi,\theta,\psi]$, where $\phi$ represents the roll, $\theta$ the pitch, and $\psi$ the yaw so that the rotation matrix that transforms the body coordinates to world coordinates can be expressed as:

\begin{equation} 
R = R_z(\psi)R_y(\theta)R_x(\phi) 
\end{equation}

\noindent Note that while they are neglected in \cite{cheetahConvexMPC,WBIC}, the rotation matrix $R$ includes the rotations about the $y$- and $x$-axes ($R_y$, $R_x$ respectively) in our work as the control of the system's dynamics is more precise if the angles $\theta$ and $\phi$ are also allowed to vary.

Note that the small angular velocity approximation allows for a simplification of Eq.~\eqref{eq:mom_integral} as in \cite{cheetahConvexMPC,WBIC}, but in the present work we consider the full dynamics so that:

\begin{equation}
\frac{d}{dt}(\mathcal{I}\omega) = \mathcal{I}\dot{\omega} + \omega \times \mathcal{I} \omega.
\end{equation}

Finally, the robot dynamics is modeled with the following set of equations:

\begin{equation}
\label{eq:full_dynamics}
\frac{d}{dt} 
\begin{bmatrix}
p \\
\Theta \\
v \\
\omega_B 
\end{bmatrix} = 
\begin{bmatrix}
v \\
J^{-1} \omega_B \\
\sum^{4}_{i=1}{\frac{f_i}{m}}-g \\
\mathcal{I}_B^{-1} ( R^T\sum^{4}_{i=1}{r_i \times f_i} - \omega_B \times \mathcal{I}_B \omega_B )
\end{bmatrix},
\end{equation}

\noindent where the matrix $J^{-1}$ transforms the angular velocity in the body frame, $\omega_B$, into the rate of change of Euler angles and $\mathcal{I}_B$ - the moment of inertia in the body frame. 
Introducing the following variables: state, action, and lever parameters $x$, $u$, and $\vartheta$, respectively as:
\begin{align}
\label{eq:x}
x & := \begin{bmatrix}p & \theta & v & \omega_B \end{bmatrix}^T, \\
\label{eq:u}
u & := \begin{bmatrix}	f_1 & f_2 & f_3 & f_4 \end{bmatrix}^T, \\
\label{eq:q}
\vartheta & := \begin{bmatrix} r_1 &  r_2 & r_3  & r_4 \end{bmatrix}^T.
\end{align}

\Cref{eq:full_dynamics} can conveniently be rewritten in a compact form:
 
\begin{equation}
\label{eq:dynamics}
\dot{x} = f(x,\vartheta,u).
\end{equation}

\section{Methods}
\label{sec:methods}
\subsection{model predictive control}
\label{subsec:MPC}
Model-predictive control utilizes the dynamics outlined in the previous section \ref{sec:Environment}.
In this approach, the ground reaction forces are considered as the action \eqref{eq:u}.
Assuming we have the sequence of desired body states $\{x_{\text{des},i|k}\}_i^N$ \eqref{eq:x} and the planned leg tip positions $\{\vartheta_{i|k}\}_i^N$ \eqref{eq:q}, a common optimal control problem with a predictive controller implemented in a digital model is to minimize a cost function $J_{MPC}$ as follows:
\begin{align}
\label{eq:actor_MPC}
\min_{\{u_{i|k}\}^N_i} J_{MPC} &( x_0,\{x_{\text{des},i|k}\}_i^N|\{u_{i|k}\}_{i}^N) := \nonumber \\ 
  = \min_{\{u_{i|k}\}^N_i} &  \sum_{i=1}^N  \gamma^{i-1} r(\hat{x}_{i|k}, x_{\text{des},i|k},u_{i|k}), \\
\sut &  \hat{x}_{0,k} = x_0, \nonumber \\
\label{eq:phi}
& \hat{x}_{i+1|k} = \Phi(\delta,\hat{x}_{i|k},\vartheta_{i|k},u_{i|k}), \\
\label{eq:cont_const}
& C_{i|k} u_{i|k} = 0, \\
\label{eq:fric_const}
& D u_{i|k} \leq 0,
\end{align}
where $\gamma$ is the discount factor, $N$ is the horizon length, $x_0$ is the initial state, $\delta$ represents the prediction step size used in the numerical integration scheme $\Phi$, $r$ is the running cost.
\Cref{eq:cont_const,eq:fric_const} represent the constraints:

\begin{enumerate}
\item The contact schedule constraint \eqref{eq:cont_const} ensures that legs in the swing phase exert zero force, while limbs in contact are allowed to have any force. We assume that all actions must satisfy this constraint.
\item The friction cone constraint \eqref{eq:fric_const} guarantees the absence of sliding between the surface and the contacting limb. Any leg contact force $f_i$ must satisfy the following conditions:

\begin{align}
-\mu f_z \leq f_x \leq \mu f_z, \\
-\mu f_z \leq f_y \leq \mu f_z,
\end{align}
where $\mu$ represents the friction coefficient set to 0.3, the equation \eqref{eq:fric_const} is responsible for ensuring the fulfillment of these conditions.
\end{enumerate}

Each subsequent state can be obtained by knowing the previous action and state \eqref{eq:phi}. In this work, the Euler explicit scheme was employed for the known system dynamics \eqref{eq:dynamics}:

\begin{equation}
\label{eq:integrator}
\Phi(\delta,\hat{x}_{i|k},\vartheta_{i|k},u_{i|k}) = \hat{x}_{i|k} + \delta f(\hat{x}_{i|k},\vartheta_{i|k},u_{i|k}).
\end{equation}

The MPC formulation is presented in the \Cref{alg:MPC}.

\begin{algorithm}
 \caption{Model predictive control (MPC)}
 \begin{algorithmic}[1]
 \renewcommand{\algorithmicrequire}{\textbf{Input:}}
 \REQUIRE System model \cref{eq:dynamics}, sampling time $\delta$, prediction horizon $N$
  \WHILE {True}
	\STATE Receive the current body state $x_k$, the desired body state sequence $\{x_{\text{des},i|k}\}_i^N$, the planned leg tip positions $\{\vartheta_{i|k}\}_i^N$, and the contact schedule matrices $\{C_{i|k}\}_i^N$
	\STATE Update actor: \\ $\displaystyle \{u^*_{i|k}\}^N_i \la \min_{\{u_{i|k}\}^N_i}{J_{MPC}( x_k,\{x_{\text{des},i|k}\}_i^N|\{u_{i|k}\}_{i}^N)}$, subject to action constraints (see \cref{eq:cont_const,eq:fric_const}).The state sequence $\{\hat{x}_{i|k}\}_i^N$ is predicted via, e.g., \cref{eq:integrator}
	\STATE Apply the first action from the action sequence, namely, $u^*_{1|k}$, to the system
  \ENDWHILE
 \end{algorithmic} 
 \label{alg:MPC}
\end{algorithm}

\subsection{Roll-Out Q-Learning}
\label{subsec:RQL}
The value iteration Q-learning utilized in RQL can be represented as follows:

\begin{align}
u_k & \la \min_u{\hat{Q}(x_k,x_{\text{des},k},u;w_k)}, \\
\label{eq:w_update_init}
w_k & \la \min_v \frac{1}{2}(\hat{Q}(x_k,x_{\text{des},k},u_k;w)  - \nonumber \\
& \hat{Q}(x_{k-1},x_{\text{des},k-1},u_{k-1};w_\text{prev}) - r(x_k,x_{\text{des},k},u_k))^2,
\end{align}
where $w_\text{prev}$ is the vector of weights from the previous time step. 
For a buffer size of $M$ the weights update formula \eqref{eq:w_update_init} can be rewritten as follows:

\begin{align}
\label{eq:critic_update}
w_k  \la & \min_{v}{J_k^c}, \\
J_k^c := & \frac{1}{2} \sum_{i=k}^{k+M-1}{e_i^2(w)}, \\
e_k(w)  := & \hat{Q}(x_k,x_{\text{des},k},u_k;w) - \nonumber \\
 & r(x_k,x_{\text{des},k},u_k) - \hat{Q}(x_{k+1},x_{\text{des},k+1},u_{k+1};w_\text{prev}).
\end{align}
For the following experiments, the buffer size $M$ is set to 500.
The actor updates of RQL are as follows:

\begin{align}
\min_{\{u_{i|k}\}^N_i} J_{RQL}^a ( x_0,& \{x_{\text{des},i|k}\}_i^N|\{u_{i|k}\}_{i}^N;w_k) := \\
 = \min_{\{u_{i|k}\}^N_i} \sum^{N-1}_{i=1} & \gamma^{i-1}r(\hat{x}_{i|k}, x_{\text{des},i|k},u_{i|k}) + \\
 & \hat{Q}(\hat{x_{N|k}},x_{\text{des},N|k},u_{N|k};w_k), \\
\sut  \hat{x}_{0,k} & = x_0, \nonumber \\
\hat{x}_{i+1|k} & = \Phi(\delta,\hat{x}_{i|k},\vartheta_{i|k},u_{i|k}), \\
C_{i|k} u_{i|k} & = 0, \\
D u_{i|k} & \leq 0,
\end{align}

\noindent where all the components, except for the Q-function model, are the same as for MPC (see \cref{eq:actor_MPC}). 
The $w_k$-weights of the Q-function on the $k^{th}$ time step were updated as stated in \cref{eq:critic_update}.

\Cref{alg:RQL} gives the final realization of RQL.

\begin{algorithm}
 \caption{Roll-Out Q-Learning(MPC)}
 \begin{algorithmic}[1]
 \renewcommand{\algorithmicrequire}{\textbf{Input:}}
 \REQUIRE System model \cref{eq:dynamics}, sampling time $\delta$, prediction horizon $N$
  \WHILE {True}
	\STATE Receive the current body state $x_k$, the desired body state sequence $\{x_{\text{des},i|k}\}_i^N$, the planned leg tip positions $\{\vartheta_{i|k}\}_i^N$, and the contact schedule matrices $\{C_{i|k}\}_i^N$
	\STATE Push the current state-action pair into the buffer (experience replay)
	\STATE Update critic $w_k  \la \min_{v}{J_k^c}$ (see \cref{eq:critic_update})
	\STATE Update actor:  \\ $\displaystyle \{u^*_{i|k}\}^N_i \la \min_{\{u_{i|k}\}^N_i}{J_{RQL}^a( x_k,\{x_{\text{des},i|k}\}_i^N|\{u_{i|k}\}_{i}^N;w_k)}$, subject to action constraints (see \cref{eq:cont_const,eq:fric_const}).The state sequence $\{\hat{x}_{i|k}\}_i^N$ is predicted via, e.g., \cref{eq:integrator}
	\STATE Apply the first action from the action sequence, namely, $u^*_{1|k}$, to the system
  \ENDWHILE
 \end{algorithmic} 
 \label{alg:RQL}
\end{algorithm}

\begin{figure}[t!]
\includegraphics[width=1\textwidth]{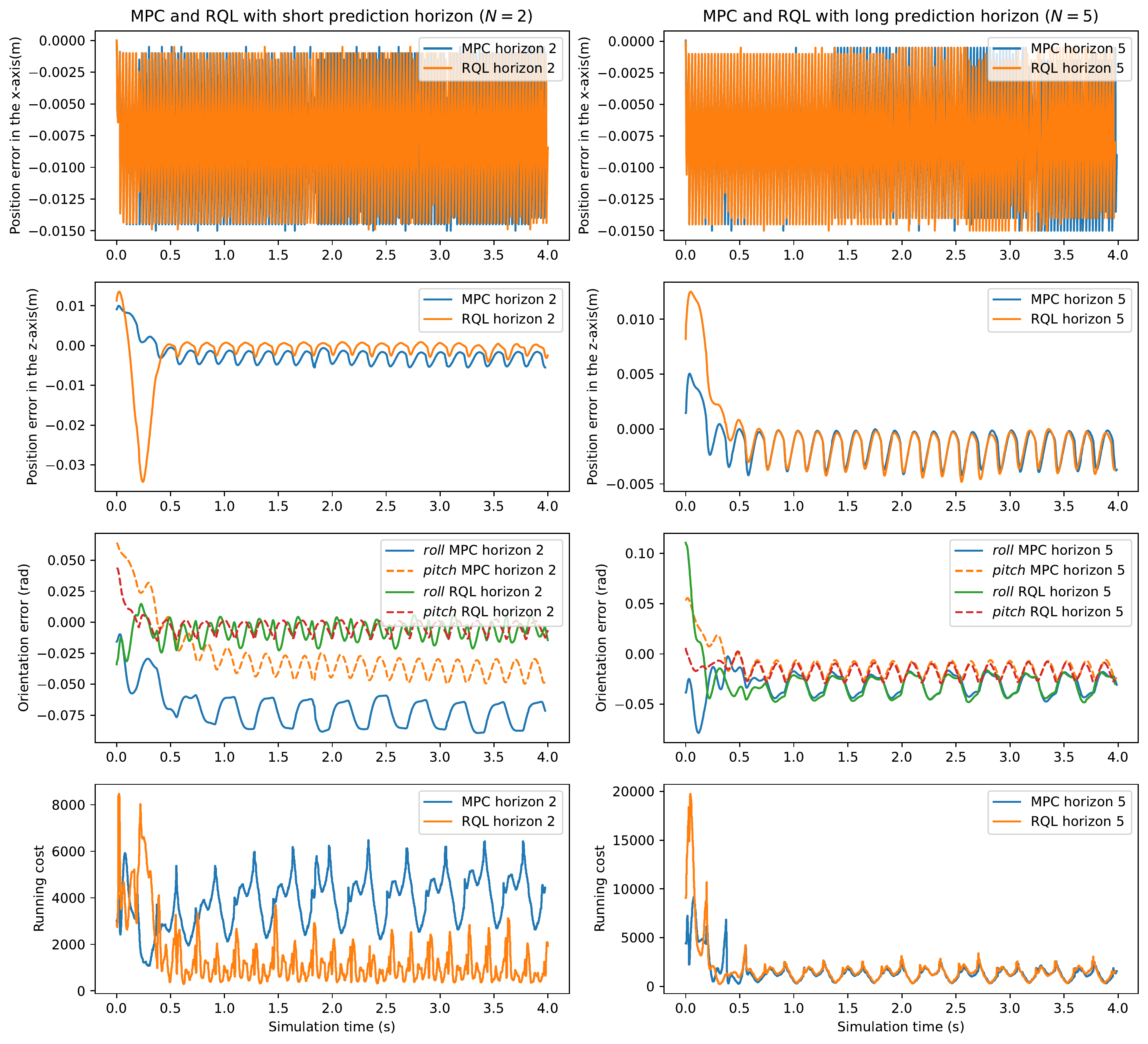}
\centering
\caption{ \textbf{Comparison of MPC and RQL with short and long prediction horizons.}\label{fig:error_res}}
\end{figure}
\section{Experimental setup}
\label{sec:experiment}

The entire experiment is built upon three frameworks: rcognita\cite{Osinenko2019rcognitaframew}, Robot Operating System (ROS), and Quad-SDK \cite{Quad-SDK}.
Rcognita is a Python package employed for the hybrid simulation of RL agents. Quad-SDK serves as a framework for the development of quadrupedal robots and acts as a planner and a simulator for the A1 Unitree robot (see \Cref{fig:robot_sim}). ROS is used to establish the connection between rcognita, functioning as a controller, and Quad-SDK.
\subsection{Method evaluation}

The running cost $r$ is defined as follows:
\begin{figure}[t!]
\includegraphics[width=0.8\textwidth]{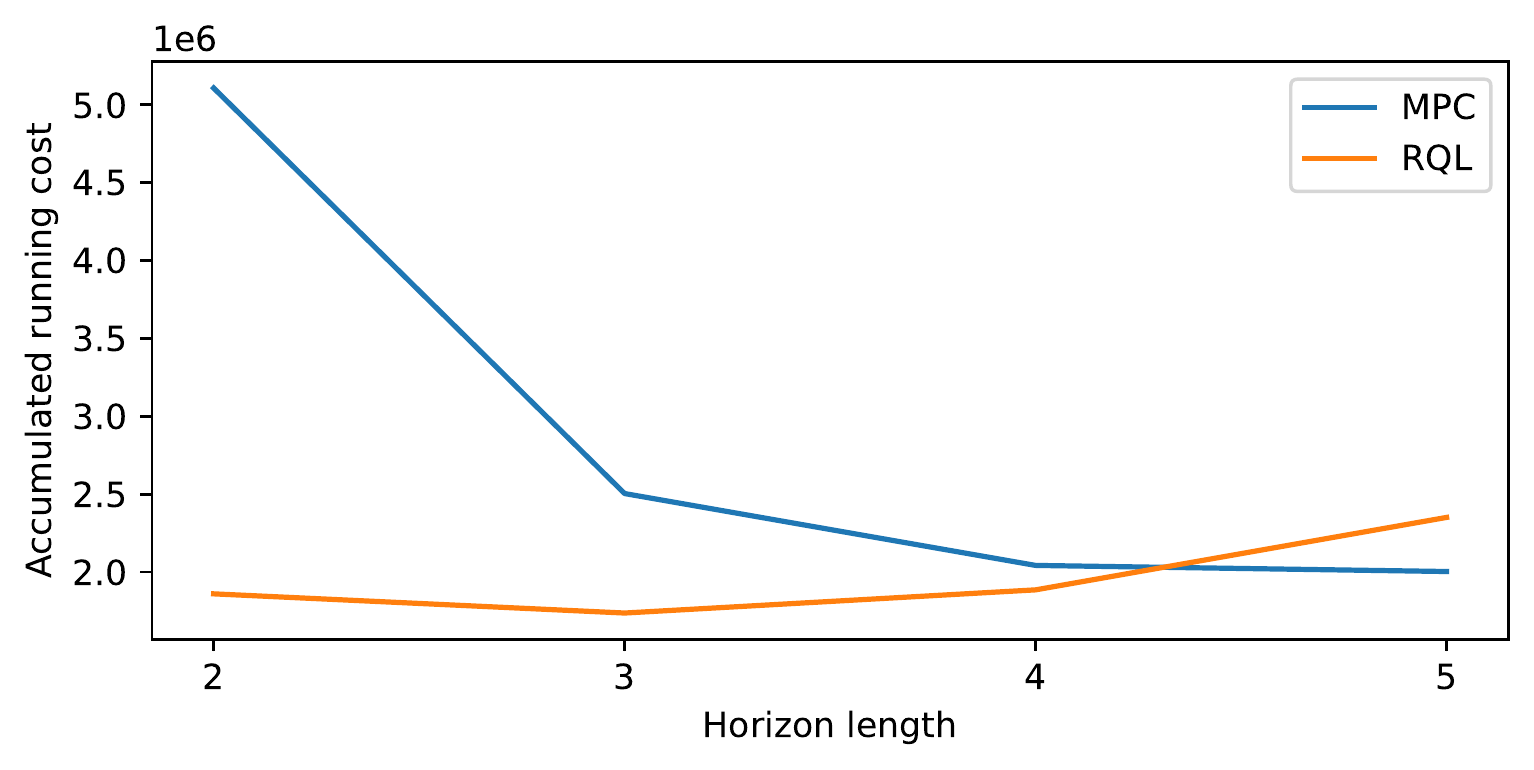}
\centering
\caption{ \textbf{Correlation between the accumulated running cost and the prediction horizon length.}\label{fig:accum_run_cost}}
\end{figure}

\begin{align}
\label{eq:rho}
r (x,x_{\text{des}},u) :=e_x^T P_x e_x + e_u^T P_u e_u,
\end{align}

\noindent where $P_x$ and $P_u$ are diagonal matrices, $P_x$ with corresponding to the body pose error, and $P_u$ corresponding to the action error.
The errors are defined as $e_x := x - x_{\text{des}}$ and $e_u := u - u_{\text{des}}$, where $u_{\text{des}} := \begin{bmatrix} \frac{mg}{4} & \frac{mg}{4} & \frac{mg}{4} & \frac{mg}{4} \end{bmatrix}^T$.
These reference ground reaction forces, denoted as $u_{\text{des}}$, are sufficient to maintain the body in the standing position.
Consequently, when the robot is in the standing position, $r (x,x_{\text{des}},u)=0$.

We model the Q-function using the following expression:

\begin{align}
\hat{Q}(x_k,x_{k,\text{des}},u_k,w)& := z^T_k A z_k, \\
\label{eq:Q_model}
z_k & := \begin{bmatrix}
x_k - x_{k,\text{des}} \\
\sum_{i=1}^4{f_{i,k} - mg}
\end{bmatrix}.
\end{align}

In this formulation, the matrix $A$ is diagonal with weights $w$ on the main diagonal. This construction is motivated by the interpretation of the Q-function. As the Q-function evaluates the state and action, it should yield a zero value when the robot is standing precisely at the desired position (i.e., $x_k = x_{k,\text{des}}$ at the $k$-th time step) with the ``ideal'' forces (i.e., the sum of forces is $mg$). The simplicity of this model is intentionally designed for computational efficiency, and we anticipate that even such a basic model could have a beneficial impact as a terminal cost.

\subsection{Results}
Both methods exhibit a peak at the beginning of the simulation due to their starting positions and the time required for RQL's initial weights to be learned through experience. for both methods (see \Cref{fig:error_res}), the error along the $x$-axis is minimal, thanks to the high value in the $P_x$ matrix that corresponds to this error. In other words, the $x$-axis error is given the highest priority.
We can compare both methods in terms of the errors along the $z$-axis and the orientation.
While MPC shows high orientation errors in a short prediction horizon, RQL significantly reduces these errors (the roll error is reduced by almost 10 times).
This can be attributed to the dominant role of the Q-function in the cost function, which allows RQL to outperform MPC.
Furthermore, we can compare the two methods in terms of running cost. RQL demonstrates superior performance over MPC with a short prediction horizon, with the mean value of the running cost being almost three times lower. However, with a longer prediction horizon, the performance of the two controllers becomes almost identical.

Examining \Cref{fig:accum_run_cost}, we can compare the performance of both methods with different prediction horizons. Notably, RQL with a short prediction horizon exhibits even better performance than MPC with a long prediction horizon. As the horizon length increases, the significance of the Q-function diminishes relative to the first terms of the cost function, making the performance of both methods almost identical.

\section{Concluding remarks}
\label{sec:results}
The current experimental results are based on a simple Q-function model \eqref{eq:Q_model}. It is worth noting that the Q-function approximation needs to be nonlinear to account for the system model's high nonlinearity. Although the linear Q-function approximation yielded significant benefits with low prediction horizon values, it is expected that a more flexible and nonlinear Q-function model would further enhance results, even for high prediction horizon values.

\bibliographystyle{IEEEtran}
\bibliography{bib/AIDA}

\end{document}